\documentclass[letterpaper, 10 pt, conference]{ieeeconf}  % Comment this line out if you need a4paper

\IEEEoverridecommandlockouts                              % This command is only needed if 
\usepackage{cite}
\usepackage{bm}
\usepackage{amsmath,amssymb,amsfonts}
\usepackage{algorithmic}
\usepackage{graphicx}
\usepackage{textcomp}
\usepackage{xcolor}
\usepackage{subfigure}
\usepackage{threeparttable}
\usepackage{multirow}
\usepackage{makecell}
\usepackage{array}  % p{}<{\centering}
\usepackage{float}%提供float浮动环境
\usepackage{booktabs}%提供命令\toprule、\midrule、\bottomrule
\def\BibTeX{{\rm B\kern-.05em{\sc i\kern-.025em b}\kern-.08em
    T\kern-.1667em\lower.7ex\hbox{E}\kern-.125emX}}
\usepackage[hidelinks]{hyperref}
\newcommand{\indexth}[1]{{#1}^{\textup{th}}}
\title{\LARGE \bf
Visual-LiDAR Odometry and Mapping with Monocular Scale Correction and Visual Bootstrapping
}

\author{Hanyu Cai$^{1}$, Ni Ou$^{1}$ and Junzheng Wang$^{1,*}$% <-this % stops a space
\thanks{*This work was supported by State Key Laboratory of
Intelligent Control and Decision of Complex Systems}% <-this % stops a space
\thanks{$^{1}$Hanyu Cai, Ni Ou, and Junzheng Wang are with School of Automation, Beijing Institute of Technology, Beijing, China. Hanyu: {\tt\small caihanyu4258@163.com}, Ni: {\tt\small 3120205431@bit.edu.cn}}%
\thanks{$^{*}$ Corresponding author:
        {\tt\small wangjz@bit.edu.cn}}%
}

\begin{document}

\maketitle
\thispagestyle{empty}
\pagestyle{empty}

%%%%%%%%%%%%%%%%%%%%%%%%%%%%%%%%%%%%%%%%%%%%%%%%%%%%%%%%%%%%%%%%%%%%%%%%%%%%%%%%
\begin{abstract}

This paper presents a novel visual-LiDAR odometry and mapping method with low-drift characteristics. The proposed method is based on two popular approaches, ORB-SLAM and A-LOAM, with monocular scale correction and visual-bootstrapped LiDAR poses initialization modifications. The scale corrector calculates the proportion between the depth of image keypoints recovered by triangulation and that provided by LiDAR, using an outlier rejection process for accuracy improvement. Concerning LiDAR poses initialization, the visual odometry approach gives the initial guesses of LiDAR motions for better performance. This methodology is not only applicable to high-resolution LiDAR but can also adapt to low-resolution LiDAR. To evaluate the proposed SLAM system's robustness and accuracy, we conducted experiments on the KITTI Odometry and S3E datasets. Experimental results illustrate that our method significantly outperforms standalone ORB-SLAM2 and A-LOAM. Furthermore, regarding the accuracy of visual odometry with scale correction, our method performs similarly to the stereo-mode ORB-SLAM2.

\end{abstract}

%%%%%%%%%%%%%%%%%%%%%%%%%%%%%%%%%%%%%%%%%%%%%%%%%%%%%%%%%%%%%%%%%%%%%%%%%%%%%%%%
\section{Introduction}
Simultaneous Localization and Mapping (SLAM) is an irreplaceable technique for mobile robots and autonomous vehicles, providing reliable surrounding environment information and real-time positions. According to the use of sensors, this technique can be divided into two categories: visual-based and LiDAR-based. Over the past two decades, visual SLAM has made significant strides, resulting in commercially available frameworks. Modern visual SLAM algorithms develop into two branches: feature-based and direct methods. Feature-based methods~\cite{ORB_SLAM, SOFT2} reduce the reprojection error of matched feature points (keypoints) through bundle adjustment (BA)~\cite{bundle_adjustment}. On the other hand, direct methods normally optimize the photometric error of sparse keypoints without corresponding matchings~\cite{DSO, LSD_SLAM}. The advantage of visual SLAM is rich semantic information, low cost and small size, which is an indispensable part of the field of automatic driving and AR.

In most cases, LiDAR SLAM usually outperforms visual SLAM. Most recent pure LiDAR SLAM methods are developed based on LOAM~\cite{LOAM}, a milestone LiDAR SLAM framework combined with SCAN-to-SCAN and SCAN-to-MAP registration modes. These LOAM-based techniques yield superior performance compared to the baseline LOAM, with improvements in efficiency~\cite{F-loam}, robust registration~\cite{T-loam}, motion compensation~\cite{CT-ICP} and local optimization~\cite{BALM}. 
% Similar patterns can be seen in LiDAR-inertia methods~\cite{Lego-LOAM,LIO-SAM}. Other LiDAR SLAM frameworks~\cite{Fast-LIO2, Faster-LIO} follow the success of FAST-LIO~\cite{Fast-LIO}, which is a different LiDAR SLAM framework based on Iterative Extended Kalman Filter with high efficiency. 
In addition, LiDAR-based loop closure detection techniques~\cite{ScanContext, ScanContext2} have been widely employed for place recognition and graph optimization to reduce the accumulated error of LiDAR SLAM further.

Nonetheless, standalone visual or LiDAR SLAM either has intractable drawbacks. Visual SLAM systems are prone to localization failure~\cite{ORB_SLAM} in fast motions. On the other hand, for LiDAR SLAM, motion distortion~\cite{LOAM, CT-ICP} is still a tricky problem for spinning LiDAR, and its loop closure detection is more complicated and challenging due to lack of stable features~\cite{ScanContext2}. It has been a noticeable trend to fuse visual and LiDAR SLAM to enhance the overall performance.

According to the fusion techniques, LiDAR-camera fused SLAM can be divided into three categories: LiDAR-assisted visual SLAM, vision-assisted LiDAR SLAM, and vision-LiDAR coupled SLAM. The first two means rely mainly on LiDAR or camera, and the other sensor takes the assistance role. Moreover, the last type generally utilizes both visual and LiDAR odometry in the system.

The first category tends to focus on image depth enhancement~\cite{LIMO} or combines with direct methods without estimating the depth of feature points~\cite{DVS_cam_LiDAR}. The second category has few related studies, and it often uses visual information to help LiDAR SLAM perform loop closure detection or render map texture~\cite{Loop-visual-LiDAR-slam,liang2016visual,3d_vision_assisted}. Since this category is not the research content of this paper, we will not describe it in detail. The third category is the hot field of current research, which can be subdivided into loosely coupling and tightly coupling. Loosely coupling is to cascade the two or filter the results of the two~\cite{LC-visual-LiDAR-slam},~\cite{SDV-LOAM}. Tightly coupling~\cite{TC-visual-LiDAR-slam} focuses on constructing a joint optimization problem, including vision and LiDAR factors for state estimation.

Our work is deeply related to depth enhancement. Whereas the error of depth enhancement is significant when the point cloud is sparse, and the feature points with enhanced depth may not be successfully tracked. Directly tracking projected points with high gradients is a solution, but such points cannot be tracked accurately and stably. In this study, we combine the powerful tracking ability of the feature-based method with optical flow and propose a novel scale correction method to address the monocular scale drift problem. Moreover, considering the LOAM algorithm depends on the constant velocity model, it is prone to failure in scenes with excessive acceleration or degradation. Using the results of the visual odometry to initialize the LiDAR odometry's pose can increase the LOAM performance.

The contributions of this paper are as follows:
\begin{enumerate}
    \item A visual-LiDAR loosely coupled odometry. Solve the problem that LOAM fails in degradative scenarios, and increase the performance.
    \item A novel scale correction algorithm is proposed that does not need to enhance the depth of the visual feature point. It guarantees that the output of the visual odometry will not have a significant drift.
    \item Implement our system on a large-scale dataset and verify its effectiveness.
\end{enumerate}

This paper is organized as follows. Section~\ref{Sec.related_work} presents studies related to our work. Section~\ref{Sec.Methodology} introduces the proposed loosely coupled system and our scale correction algorithm. Section~\ref{Sec.Experiment} shows the experimental datasets and results. Finally, Section~\ref{Sec.Conclusion} demonstrates our conclusion and possible extensions to our work.

\section{Related Work}
\label{Sec.related_work}
LiDAR-camera SLAM can be broadly classified into three categories: LiDAR-assisted visual SLAM, vision-assisted LiDAR SLAM, and vision-LiDAR coupled SLAM. Note that vision-assisted LiDAR SLAM systems~\cite{liang2016visual} are not comprehensively reviewed in this paper because this system usually hinges on semantic information, which requires knowledge of image recognition that is out of our scope.
\subsection{LiDAR-assisted Visual SLAM}
LiDAR-assisted visual SLAM generally aims to utilize LiDAR's point cloud data to obtain more accurate depth information for image feature points. A typical method in this category is LIMO, where LiDAR data is directly applied to estimate the depth of feature points~\cite{LIMO}. Yuewen et al. proposed CamVox, an RGBD SLAM system combined with Livox LiDAR~\cite{Camvox}. The performance of outdoor RGBD cameras is improved by depth enhancement, and the depth of many enhanced feature points reaches 100 meters. Another approach to using LiDAR data in visual SLAM is by projecting point clouds onto images and performing the direct method on projected points~\cite{DVS_cam_LiDAR}. Reproject the projected points to the next frame image and then minimize the photometric error to solve the pose. This method does not have the error caused by depth enhancement, but it requires accurate extrinsic parameters between the camera and LiDAR. LiDAR points are too sparse compared to image pixels, and the above methods can obtain the depth of a small number of points. In order to increase the number of pixels with depth, Varuna et al. used the Gaussian process 
regression on the projected points from LiDAR to the image to improve the depth estimation~\cite{LiDAR_cam_fusion_driveless}. Within a local image patch, they use the enhanced depth pixels as a priori to predict the depth of the remaining pixels in the image patch. In addition to depth enhancement, LiDAR can also improve the robustness of visual SLAM to illumination, which is also reflected in CamVox~\cite{Camvox}. Jiawei Mo et al. proposed a method that uses LiDAR's descriptor to address the issue that visual loop closure detection is heavily affected by illumination changes~\cite{stero_LiDAR_recog}. They calculate the LiDAR point cloud into three descriptors and store them. The stereo SLAM map is also calculated as three descriptors and matched with the LiDAR descriptors. This method only relies on three-dimensional points to complete visual loop closure detection.

To summarize, depth enhancement is the most popular technique in LiDAR-assisted visual SLAM. In this paper, we propose a novel approach that can apply to the low-resolution LiDAR case, where the density of LiDAR point clouds is much lower than that of the camera images.
\subsection{Vision-LiDAR Coupled SLAM}
\label{Sec.related_work.Vision-LiDAR Coupled SLAM}
In contrast to LiDAR-assisted visual SLAM, vision-LiDAR coupled SLAM integrates both visual and LiDAR odometry modules to enhance the system's accuracy. V-LOAM is a loosely coupled system that combines visual and LiDAR odometry modules~\cite{V-LOAM}. In this study, visual odometry recovers the depth of feature points from surrounding projected LiDAR points, while LiDAR odometry leverages high-frequency camera poses to mitigate drift. However, V-LOAM still faces two significant issues: ineffective depth enhancement and non-negligible drift error on the z-axis (also remains in its baseline~\cite{LOAM}). Zikang Yuan et al. proposed SDV-LOAM~\cite{SDV-LOAM}. It tracks the high-gradient projected LiDAR points as visual odometry and employs an adaptive scan-to-map optimization method to constrain pose in all six dimensions well. By contrast, TVL-SLAM\cite{TC-visual-LiDAR-slam} does not enhance the visual odometry's depth estimation nor utilize the motion estimation from visual odometry as the LiDAR odometry's initial guess. Instead, it establishes a joint optimization problem of visual and LiDAR features, thereby establishing a tightly coupled system.

The advantage of loose coupling is that the system structure is simple and the precision is high, but the robustness is not strong due to the influence of each module. Tight coupling is generally more robust due to joint state estimation but requires more computation.

\begin{figure}[htbp]
\centering
\includegraphics[width=0.45\textwidth]{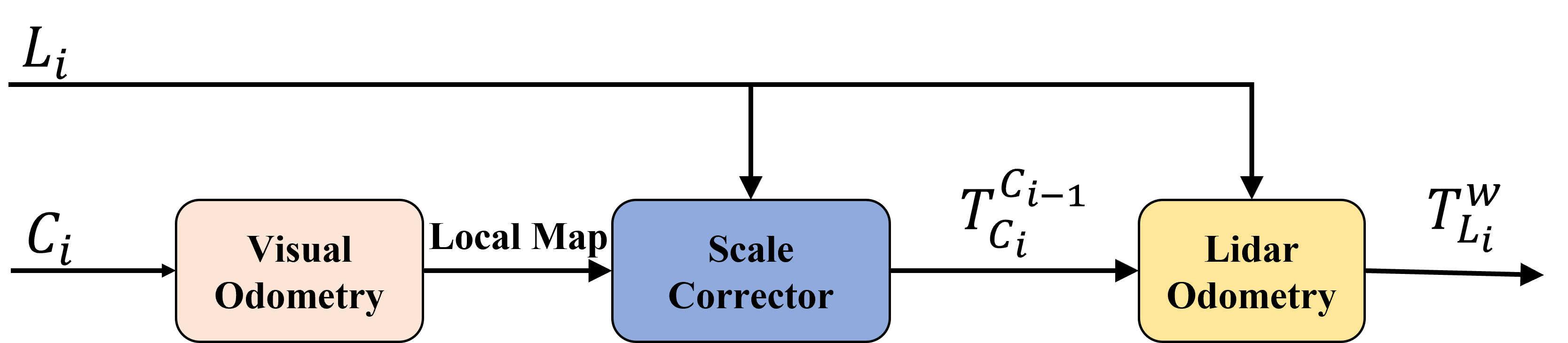}
\caption{Overview of our method.}
\label{Fig.overview}
\end{figure}
\begin{figure*}
    \centering
    \includegraphics[width=0.80\textwidth]{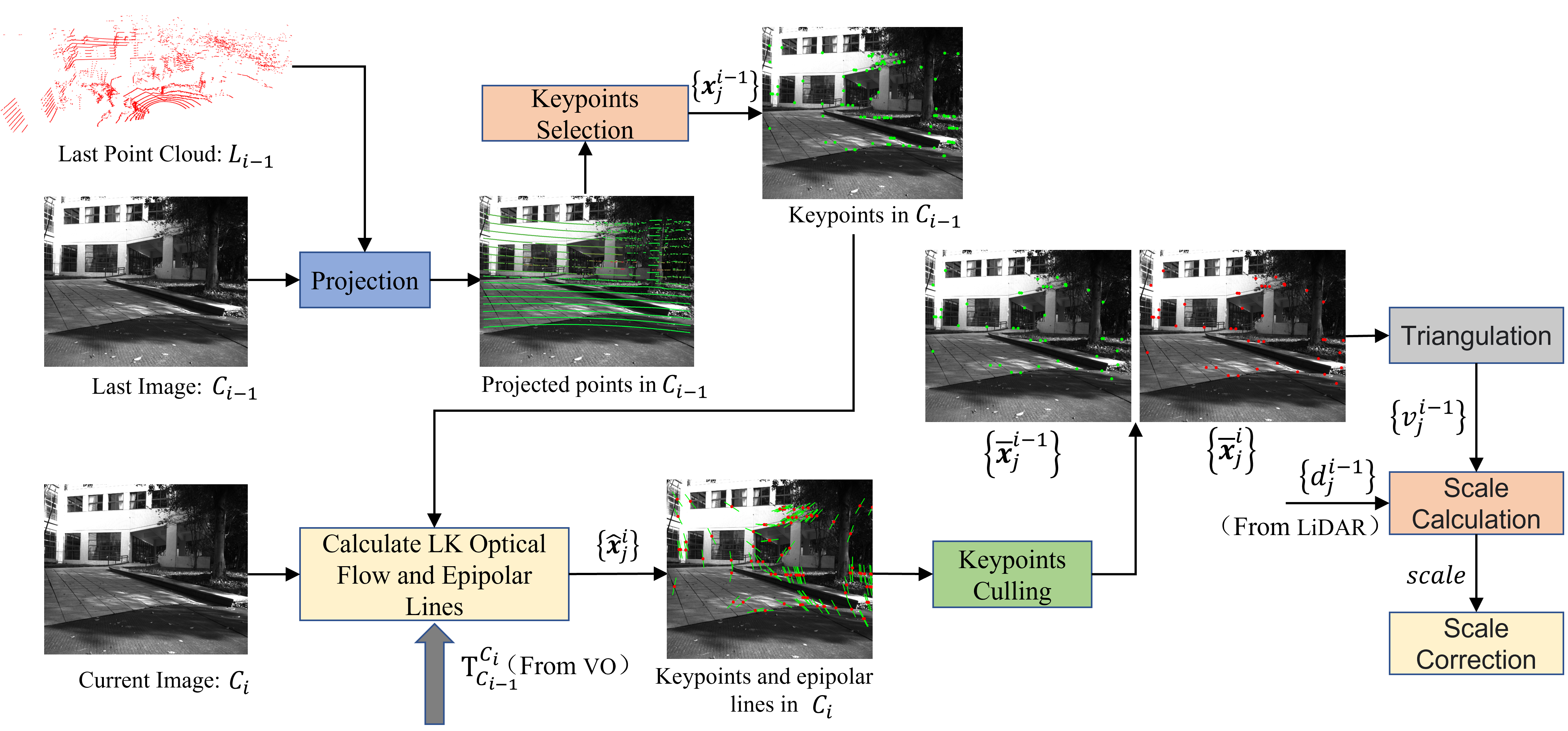}
    \caption{Pipeline of the Scale Corrector.}
    \label{scale_corrector}
\end{figure*}
\begin{table}[htbp]
    \centering
    \caption{Nomenclature}
    \begin{tabular}{p{25pt}<{\centering} p{180pt}<{\centering}}
    \toprule
    Notations & Description  \\
    \midrule
     $C_{i}$ & $\indexth{i}$ frame of image keyframe\\
     $L_{i}$ & $\indexth{i}$ frame of point cloud \\
     $\{\bm{x}_j^i\}$ &  keypoints projected from $L_{i}$ onto $C_i$ \\[1.2ex]
     $\{\widehat{\bm{x}}_j^i\}$ & reserved keypoints of $\{\bm{x}_j^i\}$ after optical flow tracking \\[1.2ex]
     $\{\overline{\bm{x}}_j^i\}$ & reserved keypoints of$\{\widehat{\bm{x}}_j^i\}$ after keypoint culling \\[1.2ex]
     $\mathbf T_{A}^{B}$ & transformation of A with respect to B \\[1.2ex]
     $\mathbf P_{w}^{i}$ & coordinates of $\indexth{i}$ map point with respect to the world \\[1.2ex]
     $\mathbf K$ & camera intrinsic matrix \\[1.2ex]
     $d_{j}^{i}$ & measured depth of the $\indexth{j}$ projected point onto $C_i$ \\[1.2ex]
     $v_{j}^{i}$ & visual depth of the $\indexth{j}$ projected point onto $C_i$ \\[1.2ex]
     $\bm{p}_j^i$ & LiDAR point corresponding to $\bm{x}_j^i$\\
    \bottomrule
    \end{tabular}
    
    \label{notation}
\end{table}
\section{Methodology}
\label{Sec.Methodology}
\subsection{System Overview}
\label{Sec.System_Overview}
The overview figure of our method is shown in Fig.~\ref{Fig.overview}, and the definitions of primary notations are present in Table \ref{notation}. Our system synchronizes the camera and LiDAR data at 10Hz. During the first stage, a local vision map is generated using the mono camera initialization or tracking. Subsequently, we utilize LiDAR data to estimate the monocular scale factor that represents the ratio between the corresponding vision local map and laser scan. However, due to the scale drift of the monocular odometry, we correct the scale factor periodically during the trajectory using the proposed scale corrector. Following scale correction, the LiDAR odometry generates the final pose with the initial guess from the visual odometry (we call it visual bootstrapping), resulting in a final localization frequency of 10 Hz. The visual odometry and LiDAR odometry are implemented based on ORB-SLAM2~\cite{ORB_SLAM} and A-LOAM~\cite{LOAM}, respectively, so we focus on performance comparison with the two baselines in our experiments part (Section~\ref{Sec.Experiment}).

The remaining parts (Section~\ref{Sec.scale_corrector_p1} and Section~\ref{Sec.scale_corrector_p2}) jointly introduce the implementation of the proposed scale corrector. The pipeline of our scale corrector is displayed in Fig.~\ref{scale_corrector}. To start with, we project the last frame of point cloud $L_{i-1}$ onto the corresponding image $C_{i-1}$ and select keypoints $\{\bm{x}_j^{i-1}\}$ among the projected points. Subsequently, the optical flow algorithm is employed to track each $\bm{x}_j^{i-1}$ in the current image $C_{i}$ and thus get $\{\widehat{\bm{x}}_j^{i-1}\}$ and $\{\widehat{\bm{x}}_j^i\}$ simultaneously. Moreover, to guarantee the accuracy of keypoint correspondence, we also design two criteria for keypoints culling based on epipolar lines, which are further introduced in (\ref{Eq.keypoint_culling_c1}) and (\ref{Eq.keypoint_culling_c2}). Based on this keypoint matching, we can conduct triangulation between matched $\{\overline{\bm{x}}_j^{i-1}\}$ and $\{\overline{\bm{x}}_j^i\}$ to recover their depth in the local map. Finally, scale correction is performed between the local map and the corresponding laser scan periodically throughout the trajectory.

\subsection{Scale Corrector: Keypoint Extraction}
\label{Sec.scale_corrector_p1}
\subsubsection{Projection and Matching}
\label{Sec.projection_matching}
As outlined in Section~\ref{Sec.System_Overview}, the content of this section includes the projection, matching and culling steps of keypoints. For clarity, we did not take image distortion into account. Then, the process of projection between $C_{i-1}$ and $L_{i-1}$ can be formulated in (\ref{Eq.projection}).
\begin{equation}
\label{Eq.projection}
\bm{x}_j^{i-1}=\frac{1}{d_j^{i-1}}\mathbf{K}\mathbf{T}_{L}^{C}\bm{p}_j^{i-1}
\end{equation}
where $\bm{p}_j^{i-1}$ is the $j^{\textup{th}}$ point of $L_{i-1}$, $\mathbf{T}_{L}^{C}$ is the extrinsic parameter between camera and LiDAR. Imprecise extrinsic will cause a significant error, and the corresponding calibration method is shown in our previous work~\cite{Targeless}.

Further, the following criteria are applied to filter out distinctive $\{{\bm{x}}_j^{i-1}\}$ through neighbouring image information.
\begin{itemize}[\noindent]
\item[a)] $\bm{x}_j^{i-1}$ should meet the requirements of the FAST-9~\cite{Fast_Corner} corner.
\item[b)] The image gradient at $\bm{x}_j^{i-1}$ should be large enough.
\end{itemize}

\begin{figure}[htbp]
    \centering
    \includegraphics[width=0.10\textwidth]{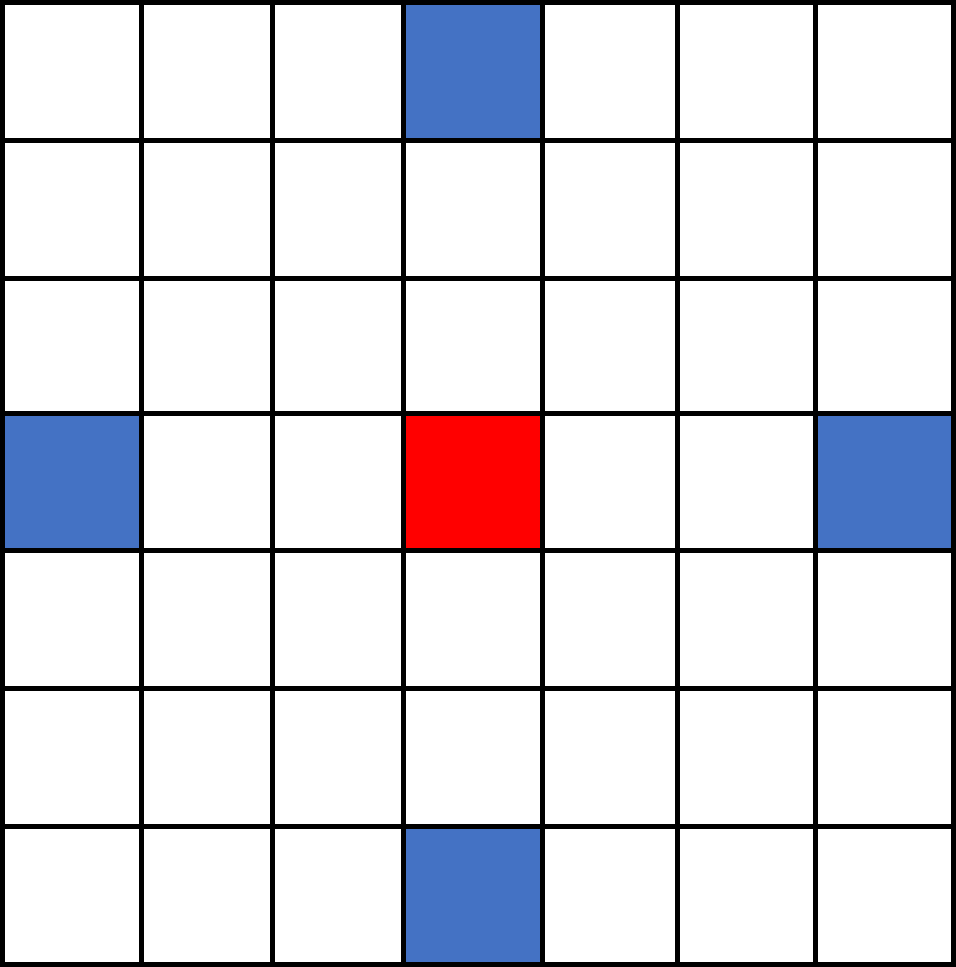}
    \caption{FAST-12 pre-testing process. The red point is the keypoint to be tested. The blue points are domain image points. }
    \label{fast12}
\end{figure}
However, the first criterion is not applicable to low-resolution LiDAR due to the scarcity of projected points. To resolve this issue, we lower the requirement to obtain $\{{\bm{x}}_j^{i-1}\}$ as shown in Fig.~\ref{fast12}. We adopt the FAST-12 pre-testing process. Calculate the difference in the pixel values between the keypoint and the surrounding four points, and if more than three meet the threshold, our requirements are met. In addition, we employ non-maximum suppression to ensure a uniform distribution of keypoints.

Regarding keypoint matching, we employ the Lucas-Kanade~\cite{LK_OP} optical flow with the input of $\{{\bm{x}}_j^{i-1}\}$ from the last image to track corresponding points $\{\widehat{\bm{x}}_j^i\}$ in the current image. 
\begin{figure}[ht]
	\centering
 \subfigure[Epipolar line]{
		\begin{minipage}[b]{0.33\textwidth}
			\includegraphics[width=1\textwidth]{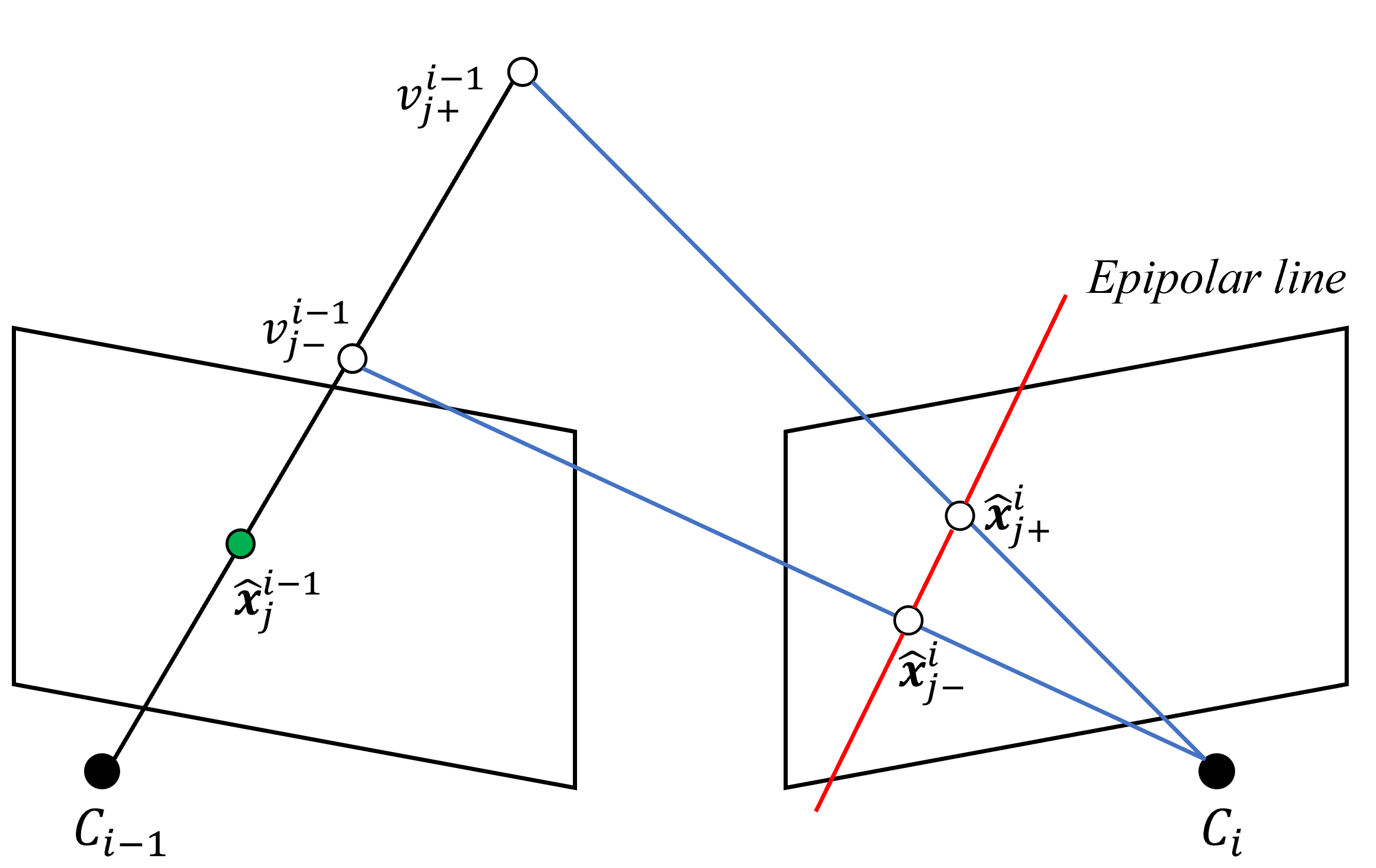}
		\end{minipage}
		\label{Epipolar line}
	}
	\subfigure[Error directions]{
		\begin{minipage}[b]{0.12\textwidth}
			\includegraphics[width=1\textwidth]{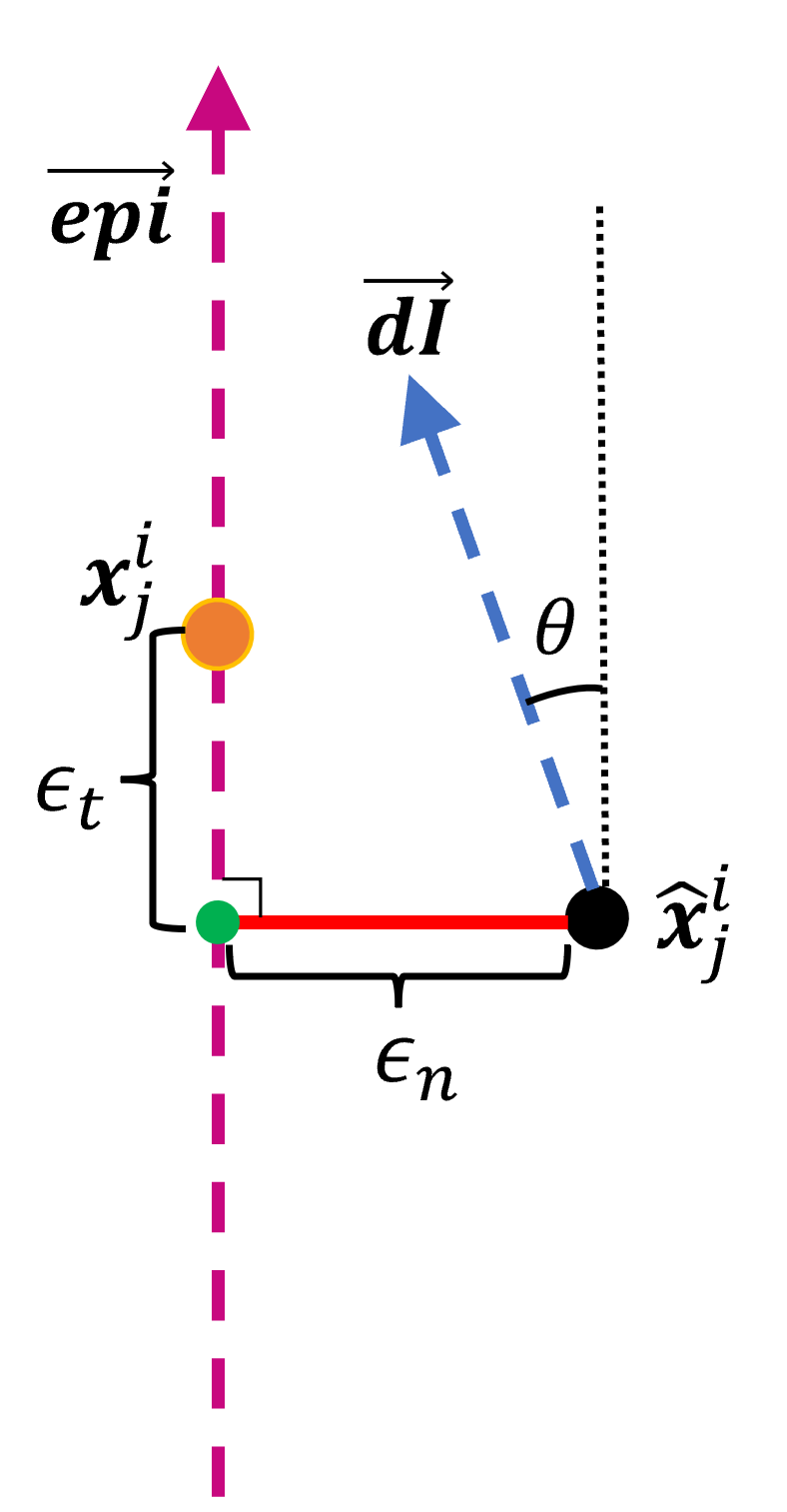}
		\end{minipage}
		\label{error_direction}
	}
	\caption{(a): The projection keypoint will theoretically lie on the epipolar line. (b): The error between the tracked keypoint(black) and the theoretical point(orange) is divided into tangential and normal errors.}
	\label{epipolar line and error}
\end{figure}
\subsubsection{Culling}
\label{Sec.culling}
Since many keypoints are not Fast corners, tracking these keypoints by optical flow will cause significant uncertainty. To further improve the accuracy of keypoint matching, we conduct keypoint culling based on the epipolar lines. Fig.~\ref{Epipolar line} shows the conception of the epipolar line. According to epipolar geometry, the keypoint $\widehat{x}_j^i$ should be located on the epipolar line. For one, it should be discarded when its distance to the epipolar line is too large. For another, in some special cases, the keypoint still should be culled when its pixel gradient is perpendicular to the epipolar line even though it is near the epipolar line. The reason is that other points distributed along the pixel gradient direction are also likely to be extracted to match this epipolar line, thereby increasing the uncertainty of its distance to the epipolar line. To cull the keypoints under the above circumstances, we propose two errors indicated in Fig.~\ref{error_direction}, denoted as normal error $\epsilon_n$ and tangential error $\epsilon_t$. We will formulate them in the following parts of this section.

According to the theory of epipolar line, $\widehat{\bm{x}}_j^{i-1}$ and $\widehat{\bm{x}}_j^i$ can theoretically be constrained by (\ref{Eq.position_range}). 
% Namely, denote $v_{j+}^{i-1}$ and $v_{j-}^{i-1}$ as the possible visual depth of $\widehat{\bm{x}}_j^{i-1}$, respectively. For each $\widehat{\bm{x}}_j^{i-1}$, the corresponding two pixel coordinates on the epipolar line can be solved through (\ref{Eq.position_range}).

\begin{equation}
\label{Eq.position_range}
\begin{split}
% \widehat{\bm{x}}_{j+}^{i}=\mathbf{K}\frac{1}{v_{j+}^{i}}\mathbf{T}_{C_{i-1}}^{C_{i}}v_{j+}^{i-1}\mathbf{K}^{-1}\widehat{\bm{x}}_j^{i-1} \\
% \widehat{\bm{x}}_{j-}^{i}=\mathbf{K}\frac{1}{v_{j-}^{i}}\mathbf{T}_{C_{i-1}}^{C_{i}}v_{j-}^{i-1}\mathbf{K}^{-1}\widehat{\bm{x}}_j^{i-1} \\
(\widehat{\bm{x}}_j^i)^T\mathbf{K}^{-T}(\mathbf{t}_{C_{i-1}}^{C_{i}})_{\times}\mathbf{R}_{C_{i-1}}^{C_{i}}\mathbf{K}^{-1}\widehat{\bm{x}}_j^{i-1}=0
\end{split}
\end{equation}

where $\mathbf{R}_{C_{i-1}}^{C_{i}}$ and $\mathbf{t}_{C_{i-1}}^{C_{i}}$ are the rotation and translation parts of $\mathbf{T}_{C_{i-1}}^{C_{i}}$ respectively, and $(\cdot)_{\times}$ represents an antisymmetric matrix. More obviously, from the formula (\ref{Eq.position_range}), the equation of the epipolar line can be obtained as $Ax+By+C=0$.

Based on this definition, the quality of tracking points can be evaluated quantitatively. As displayed in Fig.~\ref{error_direction}, we propose two evaluation metrics of different directions. Intuitively, as formulated in (\ref{Eq.keypoint_culling_c1}), the \textbf{normal error} $\epsilon_n$ is evaluated through the distance between $\widehat{\bm{x}}_j^i$ and epipolar line. We also set a threshold (0.5) to filter out fine points subject to this condition.

\begin{figure}[htbp]
\centering{\includegraphics[width=0.3\textwidth]{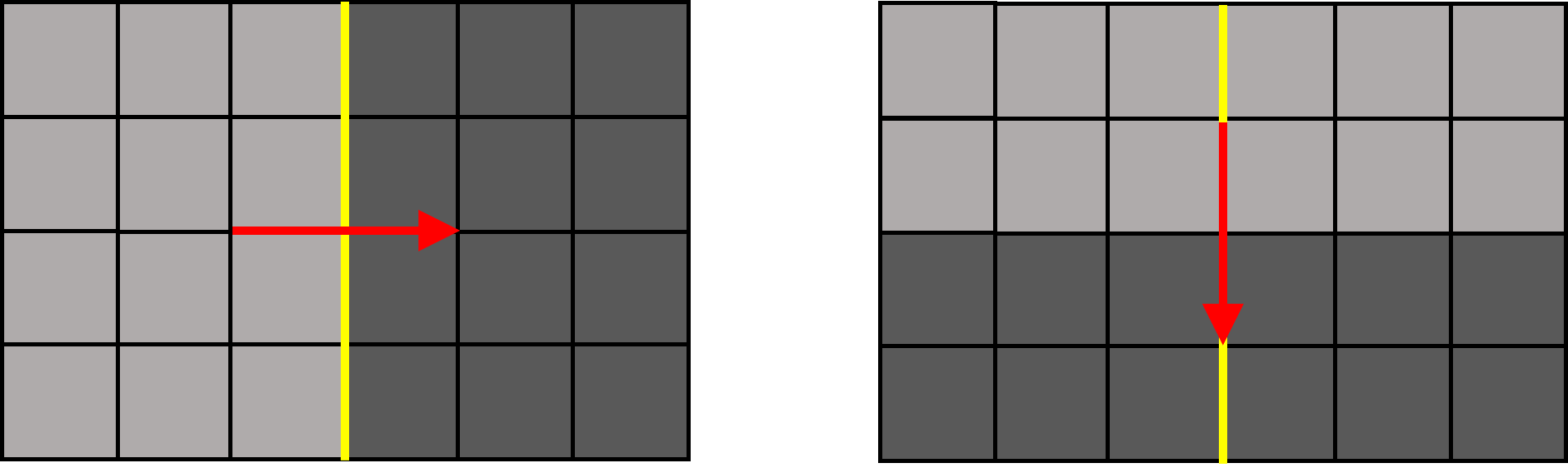}}
\caption{Two extreme cases of pixel gradient and epipolar line directions. The yellow line is the epipolar line, and the red is the pixel gradient.
\textbf{Left:} The two are perpendicular; many similar pixels are on the epipolar line. Thus, the matching uncertainty on the epipolar line is significant. \textbf{Right:} The two are parallel; the boundary pixels have a higher degree of discrimination than other pixels on the epipolar line. Thus, the matching uncertainty on the epipolar line is small.}
\label{epi2}
\end{figure}
\begin{equation}
\label{Eq.keypoint_culling_c1}
\epsilon_n=\frac{\vert A \widehat{\bm{x}}_{j.x}^i+B \widehat{\bm{x}}_{j.y}^i+C \vert}{\sqrt{A^2+B^2}} < 0.5
\end{equation}
where $\widehat{\bm{x}}_{j.x}^i$ \& $\widehat{\bm{x}}_{j.y}^i$ are the $x$ \& $y$ coordinates of $\widehat{\bm{x}}_{j}^i$, respectively.

Before explaining the tangential error, it is necessary to introduce optical flow again. Optical flow relies on pixel gradient to track the keypoint, usually using an image patch around the keypoint to increase accuracy. The same trick is used in the epipolar search~\cite{DSO}. Therefore, we can refer to the epipolar search to give a qualitative description of the tangential error. Inspired by \cite{engel2013semi}, the angle between the epipolar line direction and the pixel gradient can be used to describe the matching uncertainty along the epipolar tangential direction. Fig.~\ref{epi2} shows two extreme cases. The larger the angle between the pixel gradient and the epipolar line, the more considerable the uncertainty along the epipolar tangential direction.

Consequently, for a keypoint $\widehat{\bm{x}}_{j}^i$ tracked by optical flow, we denote $\overrightarrow{\bm{epi}}$ and $\overrightarrow{\bm{dI}}$ as the epipolar line direction vector and pixel gradient vector, respectively,  as shown in Fig.~\ref{error_direction}. Then, we can define the $|\cos{\theta} |$
% \textbf{tangential error} $\epsilon_t$ 
and its threshold in (\ref{Eq.keypoint_culling_c2}).
\begin{equation}
\label{Eq.keypoint_culling_c2}
|\cos{\theta} |=\left\vert \frac{\overrightarrow{\bm{epi}}\cdot \overrightarrow{\bm{dI}}}{\Vert \overrightarrow{\bm{epi}}\Vert \cdot \Vert \overrightarrow{\bm{dI}}\Vert}\right\vert>0.5
\end{equation}

Where $\theta$ is the angle between $\overrightarrow{\bm{epi}}$ and $\overrightarrow{\bm{dI}}$. The \textbf{tangential error} $\epsilon_t$ may be more significant if $|\cos{\theta} |$ is smaller than the threshold according to the matching uncertainty from the previous analysis.

At the end of keypoint culling, the points not subject to (\ref{Eq.keypoint_culling_c1}) and (\ref{Eq.keypoint_culling_c2}) are discarded, thereby reserving reliable matched points $\{\overline{\bm{x}}_j^i\}$ and $\{\overline{\bm{x}}_j^{i-1}\}$.

\subsubsection{Scale Calculation}
\label{Sec.scale_calculation}
With matched keypoints $\{\overline{\bm{x}}_j^i\}$ and $\{\overline{\bm{x}}_j^{i-1}\}$, we can restore the depth of each point $\overline{\bm{x}}_j^{i-1}$ by triangulation and calculate the scale factor $s_j^{i-1}$ through being dividing by the measured depth $d_j^{i-1}$, which is the distance of LiDAR point $\bm{p}_j^i$ previously projected to $C_i$ in (\ref{Eq.projection}).
\begin{equation}
\label{Eq.scale_factor}
s_j^{i-1}=\frac{d_j^{i-1}}{v_j^{i-1}}
\end{equation}

Note that there are probably a considerable proportion of outliers among $\{s_j^{i-1}\}$, so we introduce RANSAC\cite{RANSAC} for outlier rejection and output the mean of inliers as the final scale factor. 
\subsection{Scale Corrector: Scale Correction}
\label{Sec.scale_corrector_p2}
In this section, we detail how to apply scale correction to the whole SLAM system. As mentioned in Section~\ref{Sec.System_Overview}, our visual odometry is implemented based on ORB-SLAM2~\cite{ORB_SLAM}. We remove the loop closing thread and employ scale correction during local mapping process. Without loop detection and closure, the scale of local map is unstable, and thus we periodically correct the scale of local map throughout the trajectory. 

At the first stage, denote \{$\mathbf T_{w}^{C_{0}},\mathbf T_{w}^{C_{1}},\mathbf T_{w}^{C_{2}}\ldots \mathbf T_{w}^{C_{m}}$\} as the poses of keyframes in the local map and \{$\mathbf P_{w}^{0},\mathbf P_{w}^{1},\mathbf P_{w}^{2}\ldots  \mathbf P_{w}^{n}$\} as the constituent map points of the local map. Note that these values are all with respect to the world coordinate system. Therefore, we transform poses and map points to reference frame $C_0$ using $(\mathbf T_{w}^{C_{0}})^{-1}$. Subsequently, in the local map coordinate system, we can correct the scale of the local map after local bundle adjustment. Finally, the local map is transformed into the world coordinate system again for the sake of compatibility with ORB-SLAM2.

Notably, we do not frequently correct the scale, as this can interfere with the local mapping thread and cause a loss of efficiency. Instead, the scale correction is only triggered when $|scale-1|\geq 2\%$, where $scale$ is the final scale factor calculated by the scale corrector.

\begin{figure}[htbp]
	\centering
 \subfigure[S3E\_College]{
    	\begin{minipage}[b]{0.22\textwidth}
   		\includegraphics[width=1\textwidth]{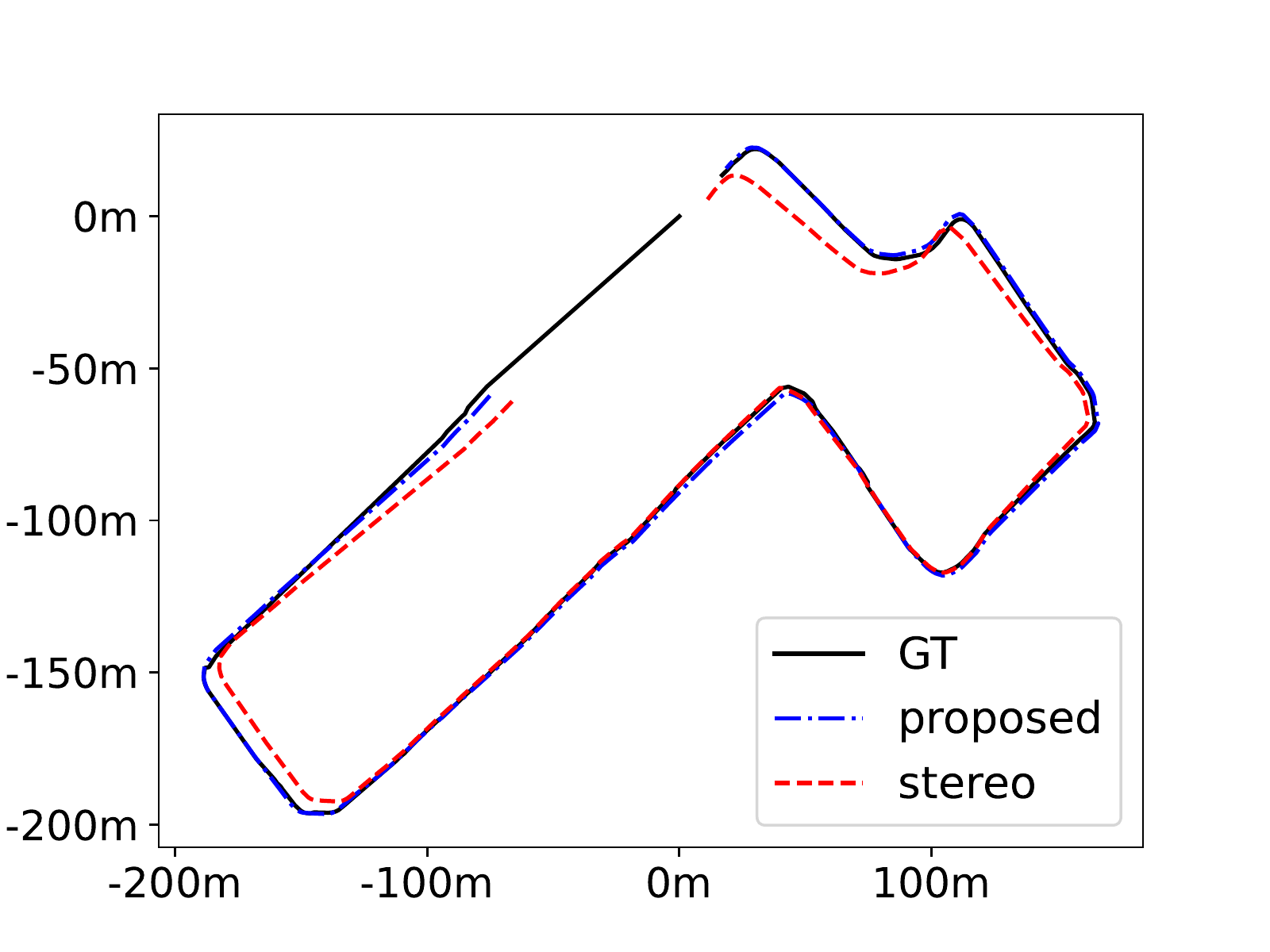}
    	\end{minipage}
	\label{traj_s3e}
    }
	\subfigure[KITTI\_00]{
		\begin{minipage}[b]{0.22\textwidth}
			\includegraphics[width=1\textwidth]{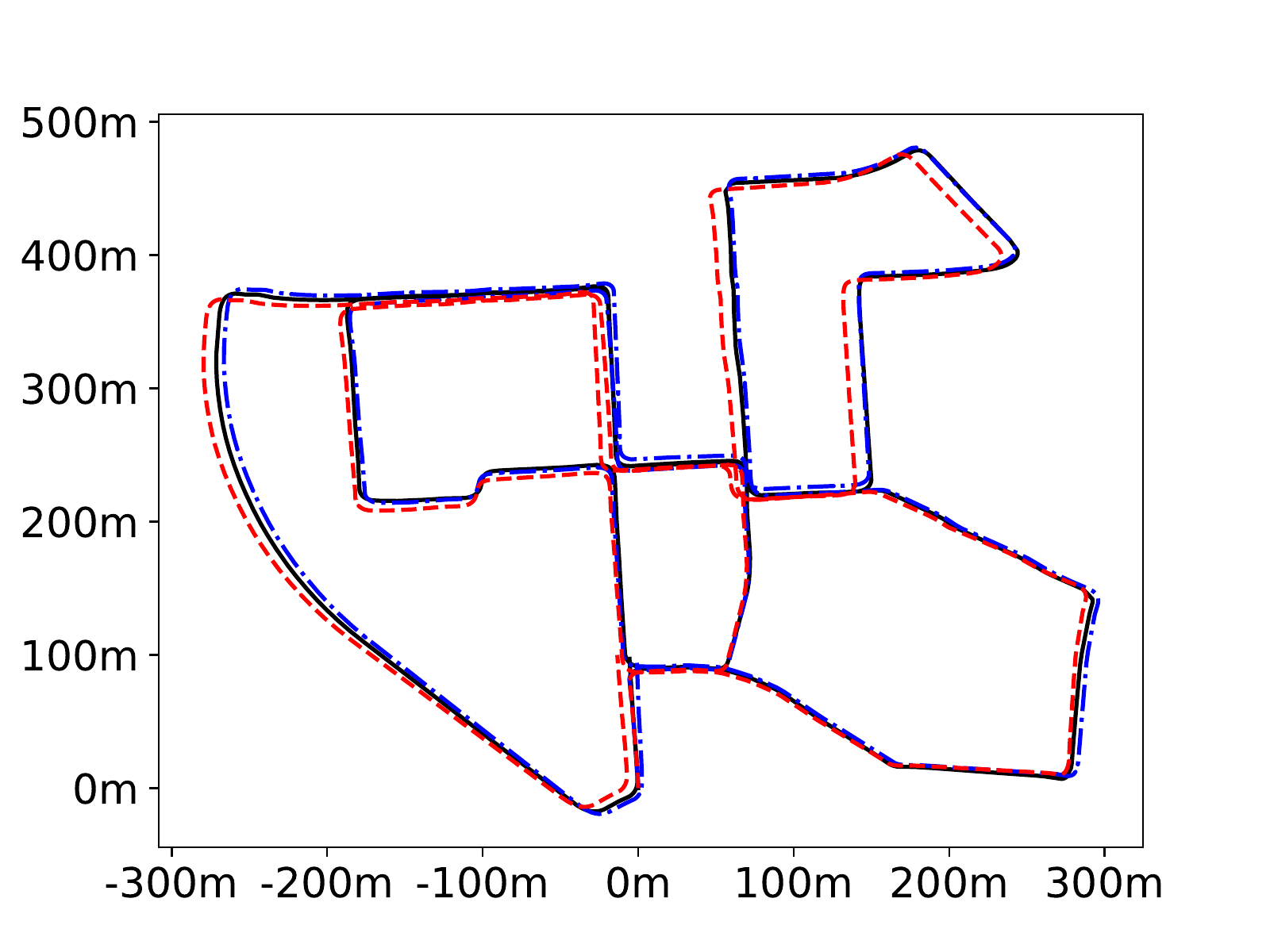}
		\end{minipage}
		\label{traj_00}
	}\\
        \subfigure[KITTI\_02]{
    	\begin{minipage}[b]{0.22\textwidth}
   		\includegraphics[width=1\textwidth]{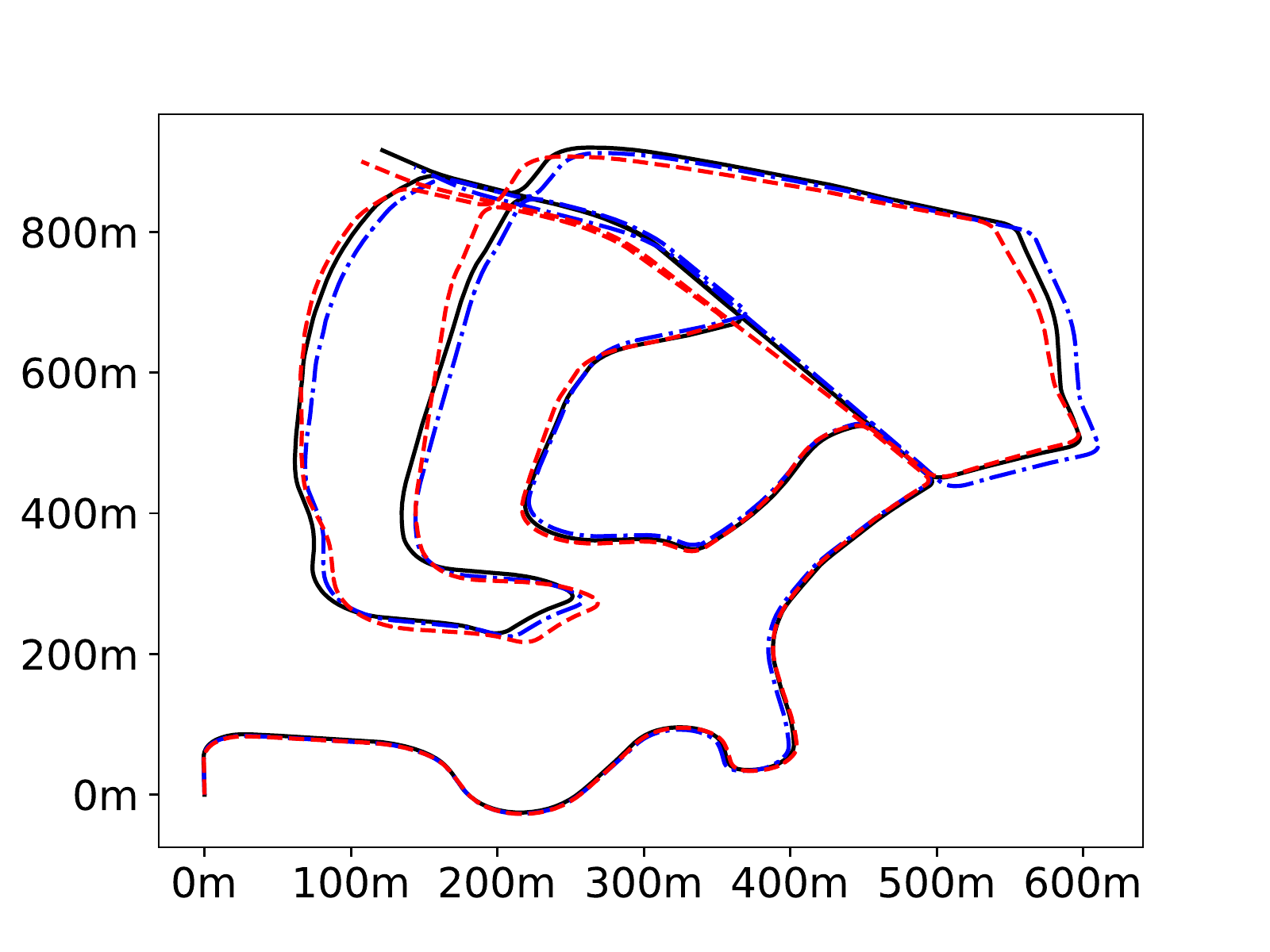}
    	\end{minipage}
	\label{traj_02}
    }
        \subfigure[KITTI\_05]{
    	\begin{minipage}[b]{0.22\textwidth}
   		\includegraphics[width=1\textwidth]{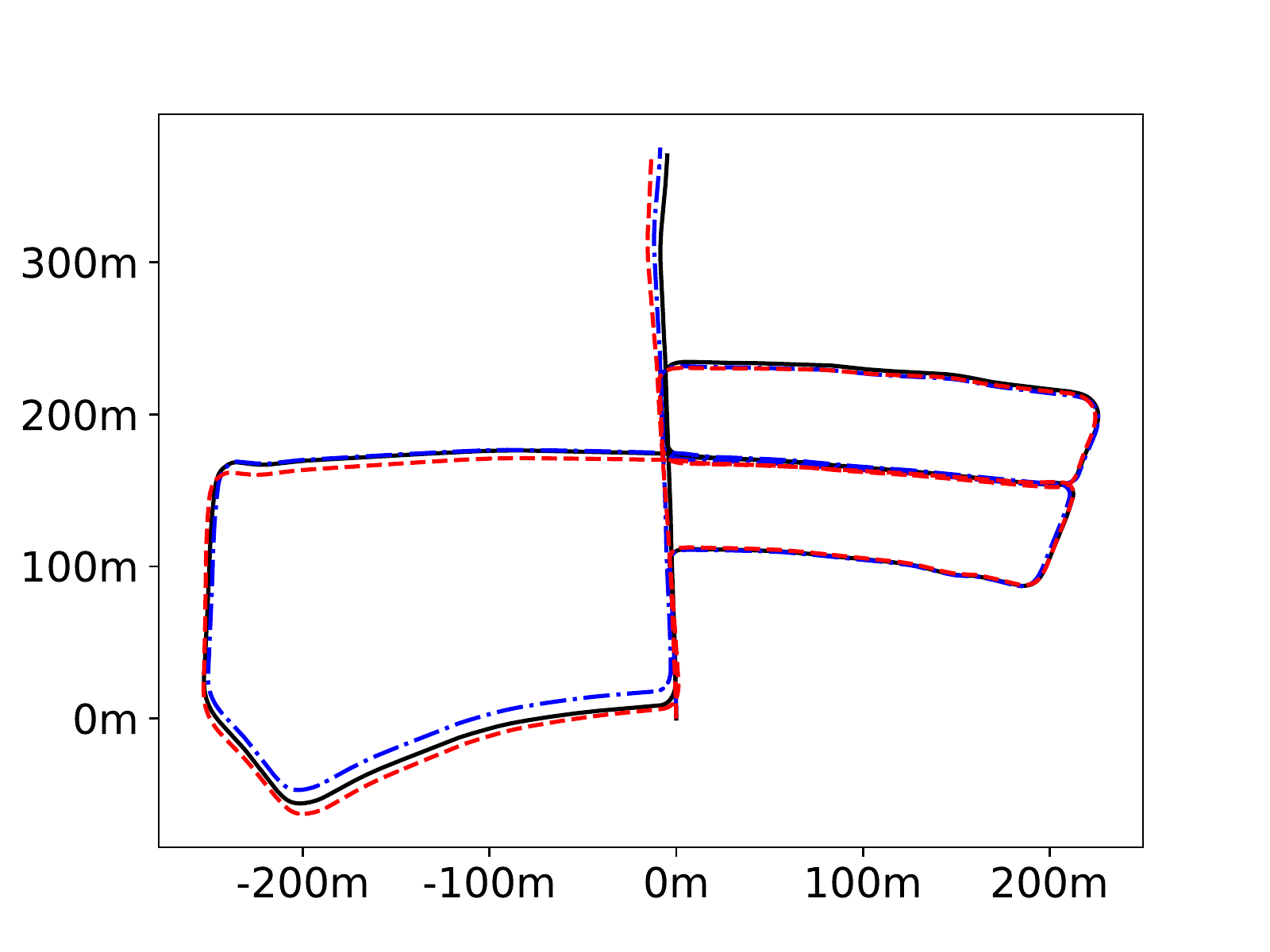}
    	\end{minipage}
	\label{traj_05}
    }\\
     \subfigure[KITTI\_08]{
    	\begin{minipage}[b]{0.22\textwidth}
   		\includegraphics[width=1\textwidth]{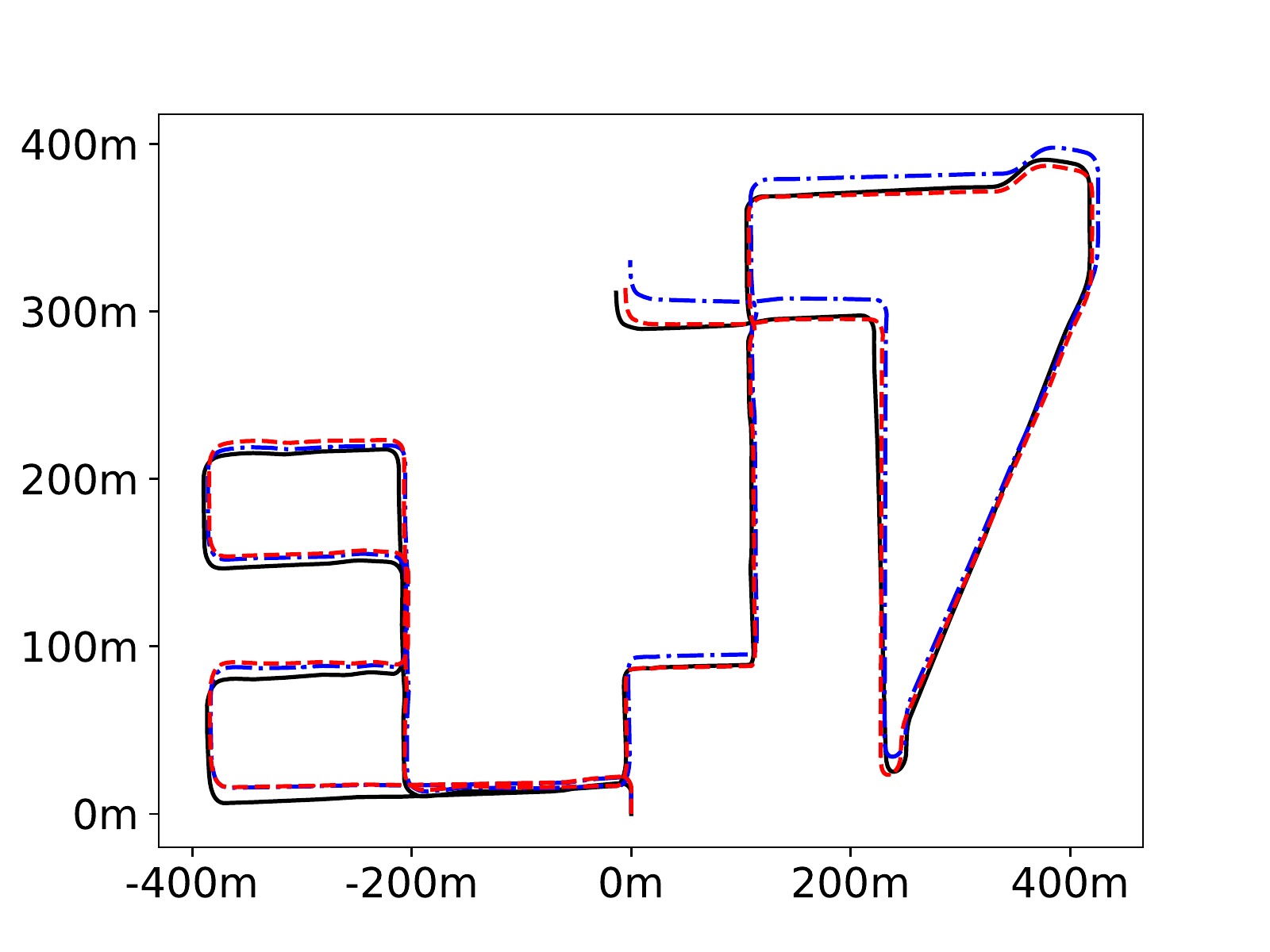}
    	\end{minipage}
	\label{traj_08}
    }

	\caption{Trajectories estimated by visual odometry. Other legends are consistent with (a). In S3E, the pose of some frames cannot be estimated due to monocular initialization.}
	\label{Trajectory diagram}
\end{figure}
\section{Experiments}
\label{Sec.Experiment}
We evaluate the performance of the proposed system on KITTI Odometry and S3E datasets. They both incorporate data collected from visual and LiDAR sensors. Four challenging sequences with long distances are selected for evaluation. Regarding data setting, KITTI Odometry uses \emph{HDL-64E} LiDAR and \emph{FL2-14S3M-C} cameras, while S3E uses \emph{VLP-16} LiDAR and \emph{HikRobot MV-CS050-10GC} cameras, which is more challenging for scale correction due to the vertical sparsity of reprojected LiDAR points. Note that we have presented a solution to the sparsity issue in Section~\ref{Sec.projection_matching}. 

% Regarding quantitative results, a popular evaluation tool evo~\cite{evo} is applied to evaluate the absolute rotation and translation error (ARE \& ATE) of predicted trajectories. 

% They are based on the relative pose error (RPE), and its definition is given in Eq.~\ref{Eq.rpe}. Subsequently, we can define RRE and RTE as shown in Eq.~\ref{Eq.rre} and Eq.~\ref{Eq.rte} respectively.

% \begin{equation}
%     \label{Eq.rpe}
%     \mathbf{RPE}_{i,j} = (\mathbf{T}_{ref,i}^{-1}\cdot \mathbf{T}_{ref,j})^{-1}\cdot(\mathbf{T}_{est,i}^{-1}\cdot \mathbf{T}_{est,j})
% \end{equation}
% \begin{equation}
%     \label{Eq.rre}
%     \mathbf{RRE}_{i,j} = \Vert log(rot(\mathbf{RPE}_{i,j})) \Vert_2
% \end{equation}
% \begin{equation}
%     \label{Eq.rte}
%     \mathbf{RTE}_{i,j} = \Vert trans(\mathbf{RPE}_{i,j}) \Vert_2
% \end{equation}

Given that our method is developed based on ORB-SLAM2~\cite{ORB_SLAM} and A-LOAM~\cite{LOAM}, we focus on comparing the localization performance of our system to that of these two baselines. In addition, we also compared with SDV-LOAM~\cite{SDV-LOAM}, one of the state-of-the-art algorithms introduced in Section~\ref{Sec.related_work.Vision-LiDAR Coupled SLAM}. All SLAM systems are performed on a laptop with a single-core AMD 6800H @3.2GHz.
% We conduct two experiments to verify the effectiveness of the method. The difference between the two experiments is the density of the LiDAR. The first experiment uses the KITTI\cite{geiger2012we} odometry dataset with 64-line LiDAR and camera data. The second experiment uses the S3E\cite{feng2022s3e} dataset, which comes with 16-line LiDAR data. Both experiments will evaluate the accuracy of visual odometry with scale correction, and the accuracy of odometry with vision loosely coupled to LiDAR.
\begin{table*}[htbp]
    \centering
    \caption{Trajectory Errors of SLAM Methods }
    \begin{threeparttable}
    \begin{tabular}{cccc|cccc}
    \Xhline{2pt}
    \rule{0pt}{10pt}
    Sequence / Length&  & Ours VO & ORB-SLAM2(Stereo)  & &Ours VLO &A-LOAM &SDV-LOAM \\[1.1ex]
    \Xhline{1pt}
    
   \multirow{2}{*}{KITTI\_00 / 3724m} &ATE(m) & \textbf{5.631} &8.946 &translationl RMSE(\%) & 1.182 &1.655 & \textbf{0.9836}    \\ 
     &ARE(deg)&\textbf{1.791 } &1.920&rotational error(deg/m)&0.0061  &0.0078 & \textbf{0.0041} \\ [1.1ex]
     
     \multirow{2}{*}{KITTI\_02 / 5067m} &ATE(m) & \textbf{13.53} &17.20  &translationl RMSE(\%) & 3.263 &11.26 & \textbf{0.8022}   \\
     &ARE(deg)&\textbf{1.821}  &3.300& rotational error(deg/m)&0.0103  &0.0307 & \textbf{0.0024} \\ [1.1ex]

     \multirow{2}{*}{KITTI\_05 / 2205m} &ATE(m) & 5.096 &\textbf{4.460} & translationl RMSE(\%) & 1.4496 &4.7189 &  \textbf{0.7036}  \\
     &ARE(deg)&\textbf{0.6319}  &1.100 &rotational error(deg/m)&0.0065  &0.0155 & \textbf{0.0030}\\ [1.1ex]

     \multirow{2}{*}{KITTI\_08 / 3222m} &ATE(m) & 13.98 &\textbf{12.47} &translationl RMSE(\%) & 1.895 &5.100  &  \textbf{1.1031}  \\
     &ARE(deg)&\textbf{1.803}  &1.824 &rotational error(deg/m)&0.0075 &0.0187&\textbf{0.0037} \\ [1.1ex]

     \multirow{2}{*}{$^{1}$S3E\_College / 920m} &ATE(m) & \textbf{1.673} &5.374 &ATE(m) & \textbf{3.097 }&5.505 & \multirow{2}{*}{$^{2}$Failed}    \\
     &ARE(deg)&--&-- &ARE(deg)&--&-- \\ [1.1ex]
     
    \Xhline{2pt}
     
    \end{tabular}
    \begin{tablenotes} 
	\item[1]	The ground truth of the S3E dataset has only the translation part, and the rotation part is the unit quaternion. 
 \item[2] SDV-LOAM fails on S3E\_College.
     \end{tablenotes}
    \end{threeparttable}
    \label{VO}
\end{table*}

\subsection{Effectiveness of Scale Corrector}
\label{Sec.Effectiveness of Scale Corrector}
To verify the effectiveness of the proposed scale corrector, we compare the absolute rotation and translation error (ATE \& ARE) between our visual odometry and the stereo-mode ORB-SLAM2. The formulation and implementation of the two metrics can be found in \textit{evo}~\cite{evo} tool. 

It should be noted that the ground-truth poses of the S3E dataset are provided by RTK without orientation (ARE is not evaluated for S3E), which worked at a much lower rate than the camera. In addition, the extrinsic calibration between RTK and camera (left) is not given. To solve these problems, we interpolate the trajectory of the visual odometry using timestamps to synchronize the predicted poses to the ground truth values using $evo$ and meanwhile employ Umeyama~\cite{umeyama} alignment between the predicted and ground-truth trajectories.
\begin{figure*}[htbp]
	\centering
 \subfigure[GT]{
		\begin{minipage}[b]{0.25\textwidth}
			\includegraphics[width=1\textwidth]{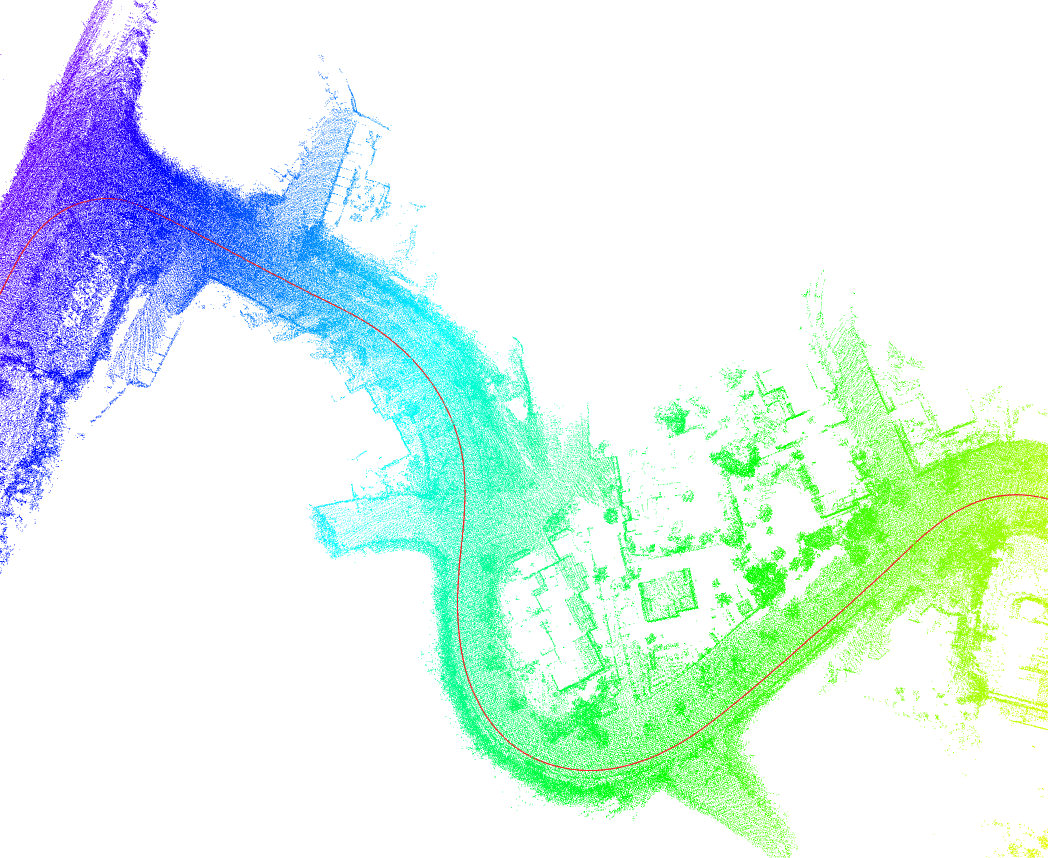}
		\end{minipage}
		\label{gt}
	}
	\subfigure[A-LOAM]{
		\begin{minipage}[b]{0.25\textwidth}
			\includegraphics[width=1\textwidth]{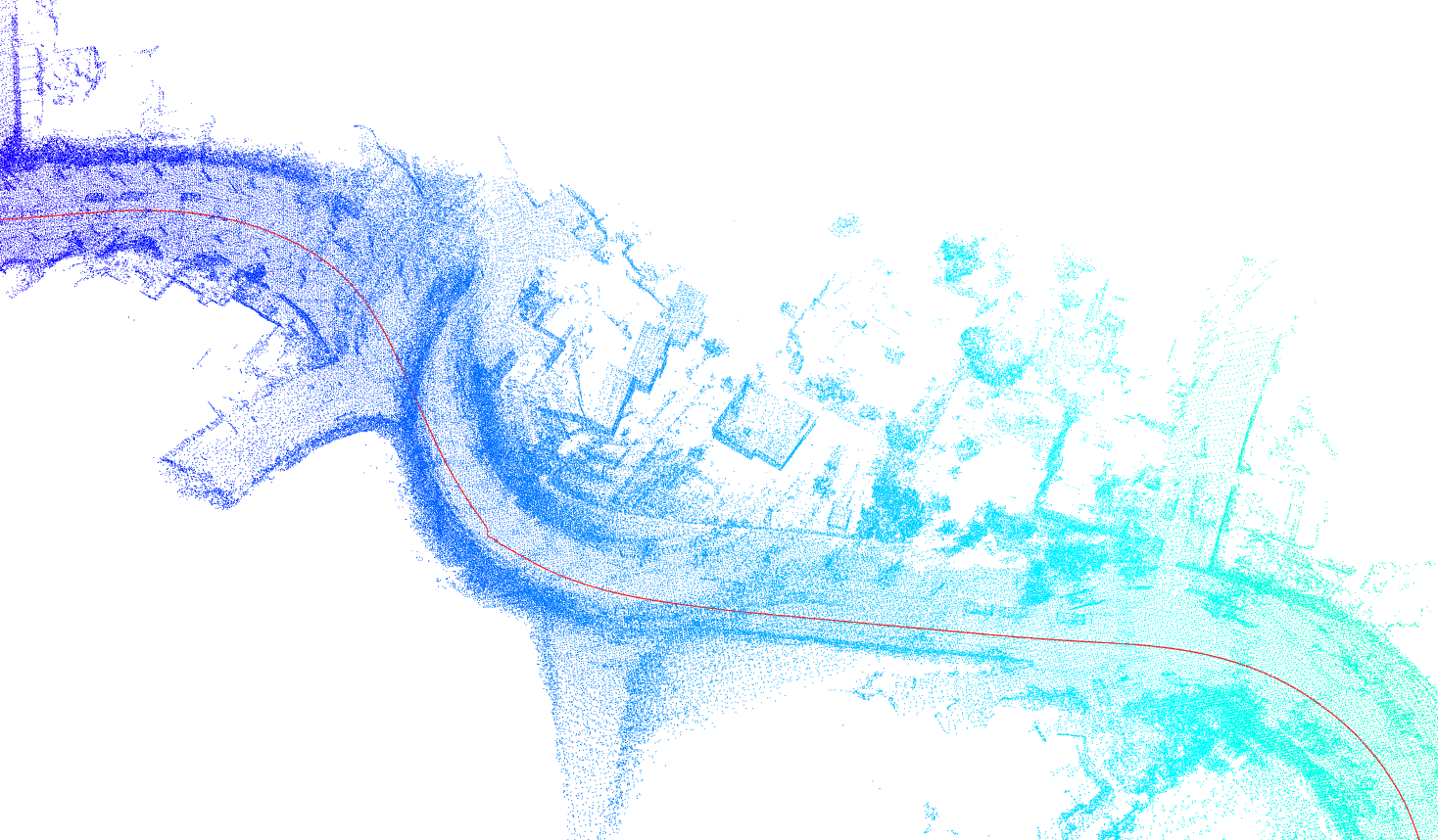}
		\end{minipage}
		\label{aloam}
	}
        \subfigure[Ours]{
    	\begin{minipage}[b]{0.25\textwidth}
   		\includegraphics[width=1\textwidth]{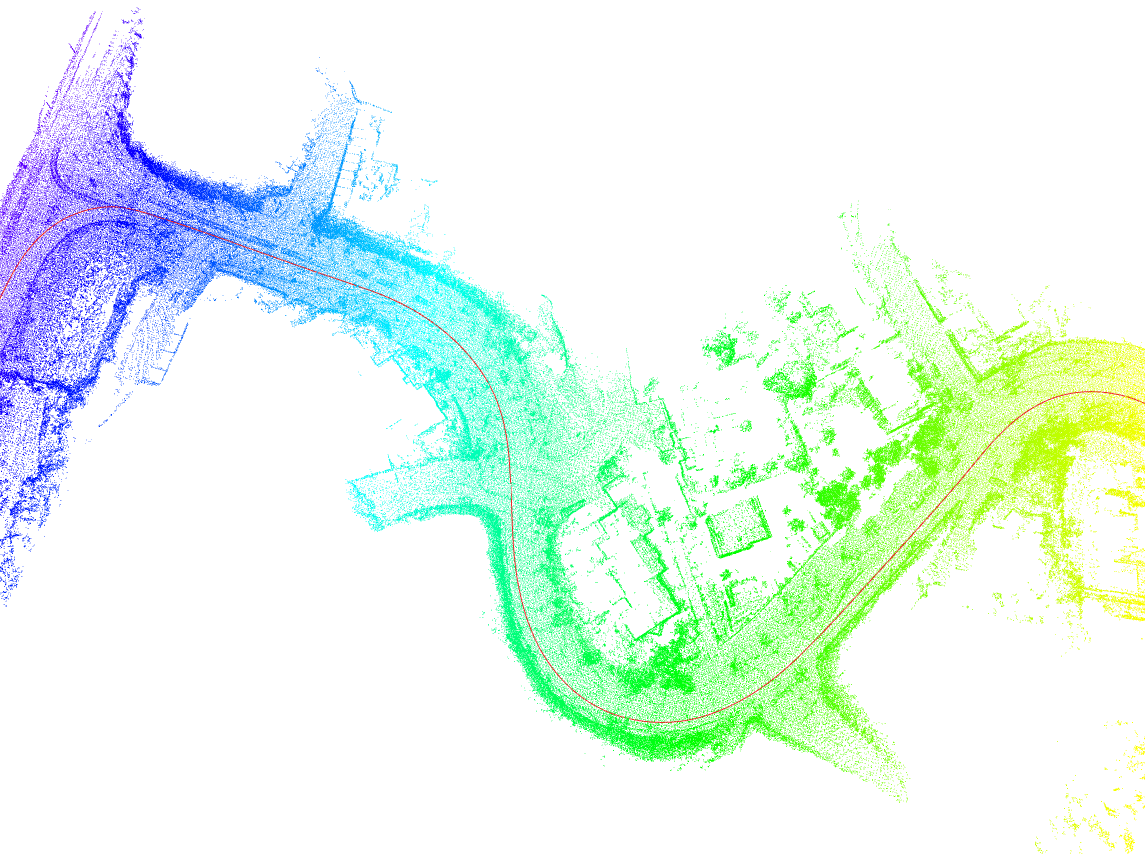}
    	\end{minipage}
	\label{ours}
    }
	\caption{Performance of degraded scenes. On a big detour with a degraded scene, A-LOAM makes wrong pose estimates, while ours works well.}
	\label{Performance of degraded scenes}
\end{figure*}
Quantitative results on five representative sequences are shown in Table~\ref{VO} while corresponding qualitative results are drawn in Fig~\ref{Trajectory diagram}. When loop closure is banned for both, our visual odometry yields better performance than stereo-mode ORB-SLAM2 in most cases, indicating the effectiveness of our scale correction module. Regarding underlying reasoning, we assume that our method is more capable of correcting the depth of distant keypoints due to the assistance of scale corrector, which is challenging for stereo vision as the parallax is not sufficient enough in this case. Moreover, we change the reference coordinate system during local optimization to the earliest keyframe in the local map, which reduce the value during optimization compared to the original solution and bring a slight performance improvement.

% \begin{table}[htbp]
%     \centering
%     \caption{Trajectory Errors of SLAM Methods  }
%     \begin{threeparttable}
%     \begin{tabular}{ccccc}
%     \toprule
%     Sequence / Length&   & Ours & A-LOAM &SDV-LOAM   \\
%     \midrule
    
%    \multirow{2}{*}{KITTI\_00 / 3724m} &translationl RMSE(\%) & \textbf{1.182} &1.655 &   \\ 
%      &rotational error(deg/m)&\textbf{0.0061}  &0.0078 & \\ [1.1ex]
     
%      \multirow{2}{*}{KITTI\_02 / 5067m} &translationl RMSE(\%) & \textbf{3.263} &11.26 &  \\
%      &rotational error(deg/m)&\textbf{0.0103}  &0.0307 & \\ [1.1ex]

%      \multirow{2}{*}{KITTI\_05 / 2205m} &translationl RMSE(\%) & \textbf{1.4496} &4.7189 &  \\
%      &rotational error(deg/m)&\textbf{0.0065}  &0.0155 & \\ [1.1ex]

%      \multirow{2}{*}{KITTI\_08 / 3222m} &translationl RMSE(\%) & \textbf{1.895} &5.100  & \\
%      &rotational error(deg/m)&\textbf{0.0075 } &0.0187&\\ [1.1ex]

%      \multirow{2}{*}{S3E\_College / 920m} &ATE(m) & \textbf{3.097 }&5.505 &  \\
%      &ARE(deg)&--&-- &\\ [1.1ex]
     
%     \bottomrule
     
%     \end{tabular}
%  %    \begin{tablenotes} 
% 	% \item[*]	The ground truth of the S3E dataset has only the translation part, and the rotation part is the unit quaternion.  
%  %     \end{tablenotes}
%     \end{threeparttable}
%     \label{VLO}
% \end{table}

\subsection{Effectiveness of Visual Bootstrapping }
\label{Sec.motion_compensation}
As for the verification of the effectiveness of Visual Bootstrapping for the LiDAR odometry, we compare it with the baseline A-LOAM~\cite{LOAM} and SDV-LOAM~\cite{SDV-LOAM} on the same datasets shown in Section~\ref{Sec.Effectiveness of Scale Corrector}. However, there is a slight difference in evaluation. For the KITTI dataset, we replace the $evo$ tool with the official KITTI evaluation tool~\cite{KITTI} for localization evaluation since it better demonstrates the drift degree in a long distance. Table~\ref{VO} illustrates that our system achieves significantly lower translation drift and slightly lower rotation drift than the A-LOAM. In the KITTI dataset, our performance is not as good as SDV-LOAM, but SDV-LOAM does not adapt to the \emph{VLP-16} LiDAR and thus fails on the S3E dataset.  

For qualitative results, we present a partial view of LiDAR map in Fig~\ref{Performance of degraded scenes}, which is part of a curved road with only trees around. In this case, A-LOAM suffers degradation while our LiDAR odometry works well. Therefore, both qualitatively and quantitatively, our method outperforms A-LOAM.
As for the reasons, A-LOAM lacks constraints on the z-axis, and the loss function easily falls into a minimum value in a degraded scene. Using the results of visual odometry to compensate for the initial value of A-LOAM can reduce the number of iterations and avoid the problem that the loss function falls into a minimum value due to the significant difference between the initial value and the actual value. 

%%%%%%%%%%%%%%%%%%%%%%%%%%%%%%%%%%%%%%%%%%%%%%%%%%%%%%%%%%%%%%%%%%%%%%%%%%%%%%%%
\section{Conclusion and future work}
\label{Sec.Conclusion}
In this study, we propose a loosely coupled monocular-LiDAR SLAM technique with a novel scale corrector. Its pose prediction derives from monocular odometry with scale correction and LiDAR odometry with visual bootstrapping. Concerning localization performance, our visual odometry achieves better performance than stereo-mode ORB-SLAM2 when loop closure for neither is available, while our LiDAR odometry significantly outperforms baseline A-LOAM~\cite{LOAM}. It is illustrated by quantitative results that the whole system yields markedly lower translation drift and moderately lower rotation drift. Qualitative results also show that our system is more robust than A-LOAM~\cite{LOAM} in degraded scenes. On the other hand, as for limitations, the proposed system relies heavily on the stability of visual odometry. In other words, a severe drift of visual odometry can cause a great loss of performance to our system, which deserves our deeper investigation. 

In our future study, we are expected to refine the proposed framework, including enhancing the robustness of visual odometry through back-end optimization, adding trouble-detection and troubleshooting tragedies for visual odometry failure and involving LiDAR points in constructing visual map.

%%%%%%%%%%%%%%%%%%%%%%%%%%%%%%%%%%%%%%%%%%%%%%%%%%%%%%%%%%%%%%%%%%%%%%%%%%%%%%%%

%%%%%%%%%%%%%%%%%%%%%%%%%%%%%%%%%%%%%%%%%%%%%%%%%%%%%%%%%%%%%%%%%%%%%%%%%%%%%%%%

%%%%%%%%%%%%%%%%%%%%%%%%%%%%%%%%%%%%%%%%%%%%%%%%%%%%%%%%%%%%%%%%%%%%%%%%%%%%%%%%

\bibliographystyle{IEEEtran}
\bibliography{ref}

\end{document}